%% file: main_pgm_avi_rl_ieee_bare_conf.tex
\documentclass[conference]{IEEEtran}

\input{bare_conf_instruction}
\renewcommand\IEEEkeywordsname{Keywords}

\input{cmd}

\usepackage{hyperref}
\begin{document}
%
\title{Tutorial and Survey on Probabilistic Graphical Model and Variational Inference in Deep Reinforcement Learning}

\author{\IEEEauthorblockN{Xudong Sun}
\IEEEauthorblockA{Department of Statistics\\Ludwig Maximillian University of Munich\\
Munich, Germany\\
Email: xudong.sun@stat.uni-muenchen.de}
\and
\IEEEauthorblockN{Bernd Bischl}
\IEEEauthorblockA{Department of Statistics\\
Ludwig Maximillian University of Munich\\
Munich, Germany
}}


\maketitle

\begin{abstract}
Aiming at a comprehensive and concise tutorial survey, recap of variational inference and reinforcement learning with Probabilistic Graphical Models are given with detailed derivations. Reviews and comparisons on recent advances in deep reinforcement learning are made from various aspects. We offer detailed derivations to a taxonomy of Probabilistic Graphical Model and Variational Inference methods in deep reinforcement learning, which serves as a complementary material on top of the original contributions.
\end{abstract}


%
\IEEEpeerreviewmaketitle
\input{sec_intro}
\input{sec_drl}

\input{sec_undirected_rbm_value_policy}

\input{sec_directed_softq}
\input{sec_vi_env}


\section{Conclusion}
As a tutorial survey, we recap Reinforcement Learning with Probabilistic Graphical Models, summarizes recent advances of Deep Reinforcement Learning and offer a taxonomy of Probabilistic Graphical Model and Variational Inference in DRL with detailed derivations which are not included in the original contributions.
\bibliographystyle{IEEEtran}
\bibliography{ref}


%
%
%

\end{document}

%% file: bare_conf_instruction.tex
\ifCLASSINFOpdf
\else
\fi

%% file: cmd.tex
\usepackage[subpreambles]{standalone}
\usepackage{amsmath}
\newcommand{\bigCI}{\mathrel{\text{\scalebox{1.07}{$\perp\mkern-10mu\perp$}}}}
\usepackage{import}

%% file: sec_intro.tex
\section{Introduction}
Despite the recent successes of Reinforcement Learning, powered by Deep Neural Networks, in complicated tasks like games \cite{mnih2015human} and robot locomotion \cite{schulman2015trust}, as well as optimization tasks like Automatic Machine Learning \cite{sun2019reinbo}. The field still faces many challenges including expressing high dimensional state and policy, exploration in sparse reward, etc. Probabilistic Graphical Model and Variational Inference offers a great tool to express a wide spectrum of trajectory distributions as well as conducting inference which can serve as a control method. 
Due to the emerging popularity, we present a comprehensive and concise tutorial survey paper with the following contributions:
\begin{itemize}
	\item We provide Probabilistic Graphical Models for many basic concepts of Reinforcement Learning, which is rarely covered in literature. We also provide Probabilistic Graphical Models to some recent works on Deep Reinforcement Learning \cite{houthooft2016vime, corneil2018efficient} which does not exist in the original contributions.
	\item We cover a taxonomy of Probabilistic Graphical Model and Variational Inference \cite{blei2017variational} methods used in Deep Reinforcement Learning and give detailed derivations to many of the critical equations, which is not given in the original contributions. Together with the recap of variational inference and deep reinforcement learning, the paper serves as a self-inclusive tutorial to both beginner and advanced readers.
\end{itemize} 

\subsection{Organization of the paper}
In section \ref{subsec:pgm_vi_basics}, we first introduce the fundamentals of Probabilistic Graphical Models and Variational Inference, then we review the basics about reinforcement learning by connecting probabilistic graphical models (PGM) in section \ref{subsec:rl_basics_pgm}, \ref{sec:vbp}, \ref{sec:pga}, as well as an overview about deep reinforcement learning, accompanied with a comparison of different methods in section \ref{sec:drl_intro}. 
In section \ref{sec:undirected}, we discuss how undirected graph could be used in modeling both the value function and the policy, which works well on high dimensional discrete state and action spaces. In section \ref{sec:softq}, we introduce the directed acyclic graph framework on how to treat the policy as posterior on actions, while adding many proofs that does not exist in the original contributions. In section \ref{sec:var_on_env}, we introduce works on how to use variational inference to approximate the environment model, while adding graphical models and proofs which does not exist in the original contributions.
\subsection{Prerequisite on Probabilistic Graphical Models and Variational Inference, Terminologies and Conventions}\label{subsec:pgm_vi_basics}We use capital letter to denote a Random Variable (RV), while using the lower case letter to represent the realization. To avoid symbol collision of using $A$ to represent advantage in many RL literature, we use $A^{act}$ explicitly to represent action. We use $(B \bigCI C) \mid A$ to represent $B$ is conditionally independent from $C$, given $A$, or equivalently $p(B|A,C) = p(B|A)$ or $p(BC|A) = P(B|A)P(C|A)$.
Directed Acyclic Graphs (DAG) \cite{bishop2006pattern} as a PGM offers an instinctive way of defining factorized joint distributions of RV by assuming the conditional independence \cite{bishop2006pattern} through d-separation \cite{bishop2006pattern}. Undirected Graph including Markov Random Fields also specifies the conditional independence with local Markov property and global Markov property \cite{asja}.

Variational Inference (VI) approximates intractable posterior distribution $p(z\mid x) = \frac{1}{\int_{z^{'}} p(z^{'})p(x|z^{'})dz^{'}}p(z)p(x\mid z)$ with latent variable $z$ specified in a probabilistic graphical model, by a variational proposal posterior distribution $q_{\phi}(z\mid x)$, characterized by variational parameter $\phi$. By optimizing the Evidence Lower Bound (ELBO) \cite{blei2017variational}, VI assigns the values to observed and latent variables at the same time. VI is widely used in Deep Learning Community like variational resampling \cite{sun2019resampling}. VI is also used in approximating the posterior on the weights distribution of neural networks for Thompson Sampling to tackle the exploration-exploitation trade off in bandit problems \cite{blundell2015weight}, as well as approximating on the activations distribution like Variational AutoEncoder \cite{kingma2013auto}.
\input{sec_svi}

\section{Reinforcement Learning and Deep Reinforcement Learning}
\subsection{Basics about Reinforcement Learning with graphical model}\label{subsec:rl_basics_pgm}
\begin{figure}[h!]
	\centering
	\includegraphics[scale=0.6]{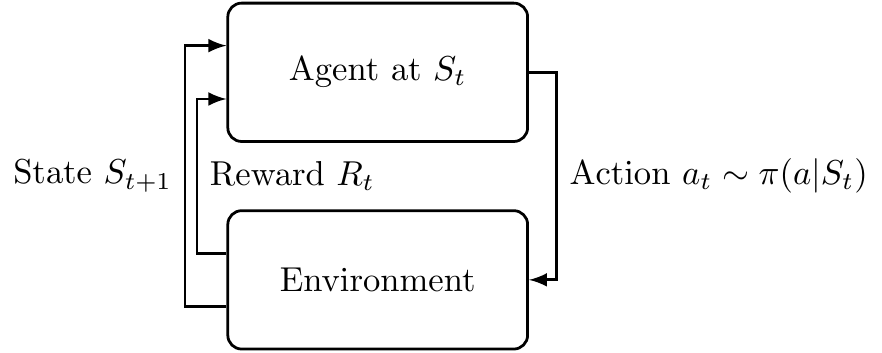}
\caption[short]{Concept of Reinforcement Learning}
\label{fig:rl}
\end{figure}

\subsubsection{RL Concepts, Terminology and Convention}\label{sec:concept}
As shown in Figure \ref{fig:rl}, Reinforcement Learning (RL) involves optimizing the behavior of an agent via interaction with the environment. At time $t$, the agent lives on state $S_t$, By executing an action $a_t$ according to a policy \cite{sutton1998introduction} $\pi(a|S_t)$, the agent jumps to another state $S_{t+1}$, while receiving a reward $R_t$. Let discount factor $\gamma$ decides how much the immediate reward is favored compared to longer term return, with which one could also allow tractability in infinite horizon reinforcement learning \cite{sutton1998introduction}, as well as reducing variance in Monte Carlo setting \cite{levine2018reinforcement}.
The goal is to maximize the accumulated rewards, $G = \sum_{t=0}^T\gamma^tR_{t}$ which is usually termed return in RL literature. 

For simplicity, we interchangeably use two conventions whenever convenient: Suppose an episode last from $t=0:T$, with $T\rightarrow \infty$ correspond to continuous non-episodic reinforcement learning. We use another convention of $t\in \{0,\cdots, \infty\}$ by assuming when episode ends, the agent stays at a self absorbing state with a null action, while receiving null reward.

\begin{figure}[h!]
	\centering
	\includegraphics[scale=0.6]{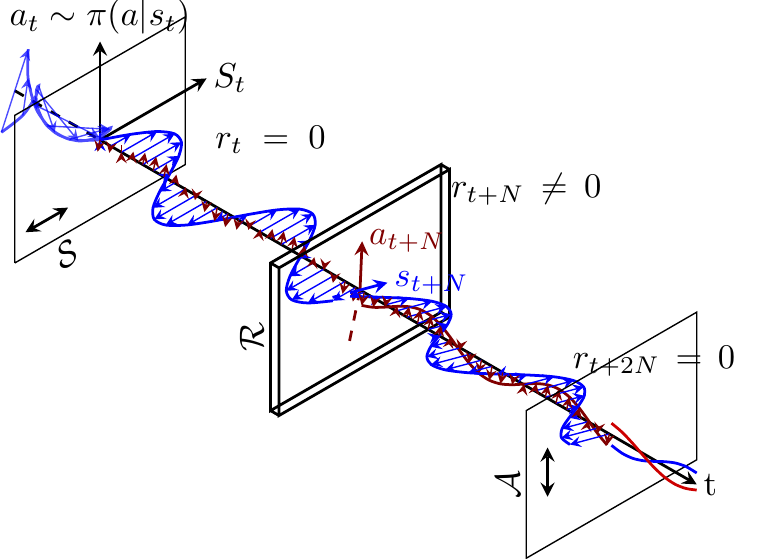}
    \caption[short]{Illustration of State, Action and Reward Trajectory}
    \label{fig:trajectory}
\end{figure}

By unrolling Figure \ref{fig:rl}, we get a sequence of state, action and reward tuples $\{(S_{t}, A^{act}_{t}, R_{t})\}$ in an episode, which is coined trajectory $\tau$ \cite{zhao2019maximum}. Figure \ref{fig:trajectory} illustrates part of a trajectory in one rollout. The state space $\mathcal{S}$ and action space $\mathcal{A}$, which can be either discrete or continuous and multi-dimensional, are each represented with one continuous dimension in Figure \ref{fig:trajectory} and plotted in an orthogonal way with different colors, while we use the thickness of the plate to represent the reward space $\mathcal{R}$. 

%
\subsubsection{DAGs for (Partially Observed ) Markov Decision Process}
\begin{figure}[h!]
	\centering
	\includegraphics[scale=0.7]{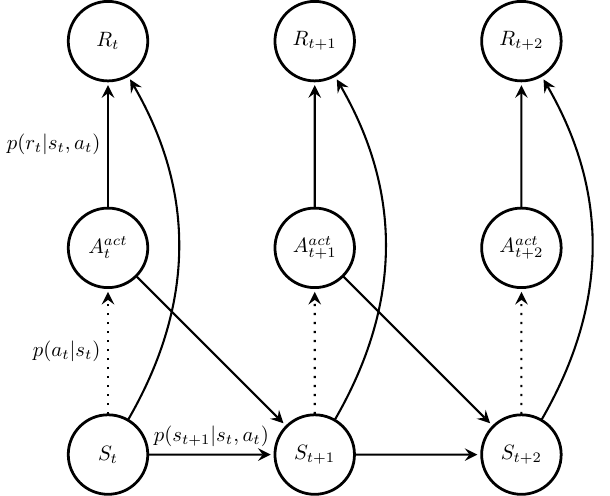}
	\caption[short]{Directed Acyclic Graph For Markov Decision Process}
	\label{fig:PGM_MDP}
\end{figure}

\begin{figure}[h!]
	\centering
	\includegraphics[scale=0.8]{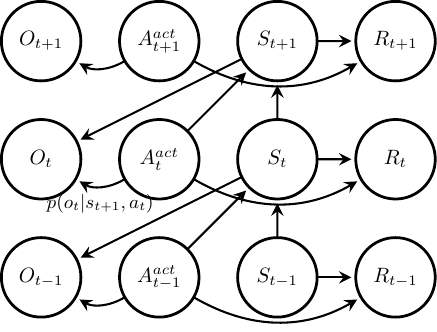}
\caption[short]{Probabilistic Graphical Model for POMDP}
\label{fig:pgm_pomdp}
\end{figure}

Reinforcement Learning is a stochastic decision process, which usually comes with three folds of uncertainty. That is, under a particular stochastic policy characterized by $\pi(a|s)=p(a|s)$, within a particular environment characterized by state transition probability $p(s_{t+1}|s_{t}, a)$ and reward distribution function $p(r_t|s_t, a_t)$, a learning agent could observe different trajectories with different unrolling realizations. This is usually modeled as a Markov Decision Process \cite{sutton1998introduction}, with its graphical model shown in Figure \ref{fig:PGM_MDP}, where we could define a joint probability distribution over the trajectory of state , action and reward RVs. In Figure \ref{fig:PGM_MDP}, we use dashed arrows connecting state and action to represent the policy, upon fixed policy $\pi$, we have the trajectory likelihood in Equation (\ref{eqn:trajectory_likelihood})
\begin{align}
p(\tau)=p(s_0)\prod_{t=0}^T p(s_{t+1}|s_{t}, a_{t})p(r_{t}|s_{t}, a_{t})\pi(a_{t}|s_{t})
\label{eqn:trajectory_likelihood}\end{align}
Upon observation of a state $s_t$ in Figure \ref{fig:PGM_MDP}, the action at the time step in question is conditionally independent with the state and action history $\mathcal{E}_t = \{S_0,A^{act}_0, \cdots, S_{t-1}\}$, which could be denoted as $(A^{act}_t\bigCI \mathcal{E}_t)\mid S_t$. 
A more realistic model, however, is the Partially Observable Markov Decision process \cite{kaelbling1998planning}, with its DAG representation shown in Figure \ref{fig:pgm_pomdp}, where the agent could only observe the state partially by observing $O_t$ through a non invertible function of the next state $S_{t+1}$ and the action $a_{t}$, as indicated the Figure by $p(o_t|s_{t+1}, a_{t})$, while the distributions on other edges are omitted since they are the same as in Figure \ref{fig:PGM_MDP}. Under the graph specification of Figure \ref{fig:pgm_pomdp}, the observable $O_t$ is no longer Markov, but depends on the whole history, however, the latent state $S_t$ is still Markov. For POMDP, belief state $b_t$ is defined at time $t$, which is associated with a probability distribution $b_t(s_t)$ over the hidden state $S_t$, with $\sum_{\mathcal{S}}b(S_t) = 1$, where state $S$ takes value in latent state space $\mathcal{S}$ \cite{kaelbling1998planning}.
\input{sec_pomdp}
\subsection{Value Function, Bellman Equation, Policy Iteration}\label{sec:vbp}
\begin{figure}[h!]
	\centering
	\includegraphics[scale=0.6]{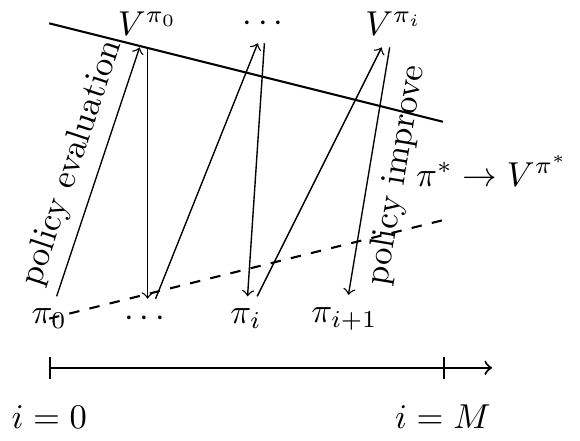}
\caption[short]{General Policy Iteration}
\label{fig:gpi}
\end{figure}

Define state value function of state $s\in \mathcal{S}$ in Equation (\ref{eqn:def_v}), where the corresponding Bellman Equation is derived in Equation (\ref{eqn:bellmanv}).
\begin{align}
&V^{\pi}(s)
= E_{\pi,\varepsilon}[\sum_{i=0}^{\infty}\gamma^i R_{t+i}(S_{t+i},  A^{act}_{t+i})\mid \forall S_t = s] \label{eqn:def_v}\\ 
=& E_{\pi,\varepsilon}[R_t(S_t, A^{act}_t)+ \gamma \sum_{i=1}^{\infty}\gamma^{(i-1)} R_{t+i}(S_{t+i}, A^{act}_{t+i})] \nonumber\\
=& E_{\pi,\varepsilon}[R_t(S_t, A^{act}_t)+ \gamma \sum_{i^{'}=0}^{\infty}\gamma^{i^{'}} R_{t+1+ i^{'}}(S_{t+1+i^{'}}, A^{act}_{t+1+i^{'}} )] \nonumber\\
=& E_{\pi,\varepsilon}[R_t(S_t, A^{act}_t)+\gamma V^{\pi}(S_{t+1})] \label{eqn:bellmanv}
\end{align}
where $S_{t+i} \sim p(s_{t+i+1}|s_{t+i}, a_{t+i})$ takes value from $\mathcal{S}$, $A^{act}_{t+i} \sim \pi(a|S_{t+i+1})$ taking value from $\mathcal{A}$, and we have used the $\pi$ and $\varepsilon$ in the subscript of the expectation $E$ operation to represent the probability distribution of the policy and the environment (including transition probability and reward probability) respectively.
State action value function \cite{sutton1998introduction} is defined in Equation (\ref{eqn:defq}), where in Equation (\ref{eqn:v2q}), its relationship to the state value function is stated.
\begin{align}
	&Q^{\pi}(s,a)\, (\forall S_t = s, A^{act}_t = a)\nonumber\\=& E_{\pi, \varepsilon}[R_t(S_t = s, A^{act}_t = a) + \sum_{i=1}^{\infty}\gamma^i R_{t+i}(S_{t+i}, A^{act}_{t+i})]\label{eqn:defq}\\
	=&E_{\pi, \varepsilon}[R_t(S_t = s, A^{act}_t = a) + \gamma V^{\pi}(S_{t+1})]\label{eqn:v2q}
\end{align}
Combining Equation (\ref{eqn:bellmanv}) and Equation (\ref{eqn:defq}), we have 
\begin{equation}V(s) = \sum_a\pi(a|s)Q(s,a)\label{eqn:q2v}
\end{equation}
Define optimal policy \cite{sutton1998introduction} to be \begin{align}
\pi^{*} &= \underset{\pi}{arg\,max}\,V^{\pi}(s),\forall s \in S \nonumber\\&=\underset{\pi}{arg\,max}\,E_{\pi}[R_t+\gamma V^{\pi}(S_{t+1})]
\end{align}
Taking the optimal policy $\pi^{*}$ into the Bellman Equation in Equation (\ref{eqn:bellmanv}), we have
\begin{align}
V^{\pi^{*}}(s)=E_{\pi^{*},\varepsilon}\left[R_t(s,A^{act}_t)+\gamma V^{\pi^{*}}(S_{t+1})\right]\label{eqn:defv_opt} 
\end{align}
 Taking the optimal policy $\pi^{*}$ into Equation (\ref{eqn:defq}), we have 
\begin{align}
	Q^{\pi^{*}}(s,a)= E_{\pi^{*}, \varepsilon}[R_t(s,a) + \sum_{i=1}^{\infty}\gamma^i R_{t+i}(S_{t+i}, A^{act}_{t+i})]\label{defq_opt}
\end{align}
Based on Equation (\ref{defq_opt}) and Equation (\ref{eqn:defv_opt}), we get 
\begin{align}
V^{\pi^{*}}(s) = \underset{a}{max}\, Q^{\pi^{*}}(s,a)
\end{align}
and 
\begin{align}
Q^{\pi^{*}}(s,a)=E_{\varepsilon, \pi^{*}}\left[R_t(s,a) + \gamma \underset{\bar{a}}{max}\,Q^{\pi^{*}}(S_{t+1},\bar{a})\right]
\end{align}
For learning the optimal policy and value function, General Policy Iteration \cite{sutton1998introduction} can be conducted, as shown in Figure \ref{fig:gpi}, where a contracting process \cite{sutton1998introduction} is drawn. Starting from initial policy $\pi_0$, the corresponding value function $V^{\pi_0}$ could be estimated, which could result in improved policy $\pi_1$ by greedy maximization over actions. The contracting process is supposed to converge to the optimal policy $\pi^{*}$. 

As theoretically fundamentals of learning algorithms, Dynamic programming and Monte Carlo learning serve as two extremities of complete knowledge of environment and complete model free \cite{sutton1998introduction}, while time difference learning \cite{sutton1998introduction} is more ubiquitously used, like a bridge connecting the two extremities. Time difference learning is based on the Bellman update error 
 $\delta_t = Q(s_t,a_t) - \left(R_t(s,a) + \gamma \underset{a}{max}\,Q(s_{t+1},a)\right)$. 
\subsection{Policy Gradient and Actor Critic}\label{sec:pga}
Reinforcement Learning could be viewed as a functional optimization process.
We could define an objective function over a policy $\pi_{\theta}(a|s)$, as a functional, characterized by parameter $\theta$, which could correspond to the neural network weights, for example. 

Suppose all episodes start from an auxiliary initial state $s_0$, which with probability $h(s)$, jumps to different state $s \in \mathcal{S}$ without reward. $h(s)$ characterizes the initial state distribution which only depends on the environment. Let $\eta(s)$ represent the expected number of steps spent on state $s$, which can be calculated by summing up the $\gamma$ discounted probability $P^{\pi}(s_0 \rightarrow s, k+1)$ of entering state $s$ with $k+1$ steps from auxiliary state $s_0$, as stated in Equation (\ref{eq:avestep}), which can be thought of as the expectation of  $\gamma^k$ conditional on state $s$. 
\begin{figure}
\begin{align}
\eta(s) &= \sum_{k=0} \gamma^kP^{\pi}(s_0 \rightarrow s, k+1)\label{eq:avestep}\\
&= h(s) + \sum_{\bar{s},a}\gamma\eta(\bar{s})\pi_{\theta}(a|\bar{s})P(s|\bar{s},a)\label{eq:avestep1}\end{align}\end{figure}
In Equation (\ref{eq:avestep1}), the quantity is calculated by either directly starting from state $s$, which correspond to $k=0$ in Equation (\ref{eq:avestep}), or entering state $s$ from state $\bar{s}$ with one step, corresponding to $k+1\geq2$ in Equation (\ref{eq:avestep}). 

For an arbitrary state $s\in \mathcal{S}$, using $s^{'}$ and $s^{''}$ to represent subsequent states as dummy index,
\begin{figure}
\begin{align}
&\nabla_{\theta}V^{\pi(\theta)}(s)=\nabla_{\theta} \left[\sum_a Q^{\pi(\theta)}(s,a)\pi_{\theta}(a|s)\right]\\
=&\sum_a\left[ \nabla_{\theta}Q^{\pi(\theta)}(s,a)\pi_{\theta}(a|s)+ \nabla_{\theta}\pi_{\theta}(a|s)Q^{\pi(\theta)}(s,a)\right] \nonumber\\
=&\sum_a \nabla_{\theta} \left[\sum_{s^{'}, R} P(s^{'}, R|s,a)\left(R + \gamma V^{\pi(\theta)}(s^{'})\right)\right]\pi_{\theta}(a|s) \nonumber\\&+ \sum_a\nabla_{\theta}\pi_{\theta}(a|s)Q^{\pi(\theta)}(s,a)\\
=&\sum_a\sum_{s^{'}}\gamma P(s^{'}|s,a)\nabla_{\theta}V^{\pi(\theta)}(s^{'})\pi_{\theta}(a|s)+\nonumber\\& \sum_a\nabla_{\theta}\pi_{\theta}(a|s)Q^{\pi(\theta)}(s,a)\label{eq:pg_org}\\
=&\sum_a\nabla_{\theta}\pi_{\theta}(a|s)Q^{\pi(\theta)}(s,a) + \sum_a\sum_{s^{'}}\gamma P(s^{'}|s,a)\pi_{\theta}(a|s)  \nonumber\\
&\left[\sum_{a^{'}}\sum_{s^{''}}\gamma P(s^{''}|s^{'},a^{'})\nabla_{\theta}V^{\pi(\theta)}(s^{''})\pi_{\theta}(a^{'}|s^{'}) + \right.\nonumber\\ &\left.\sum_{a^{'}}\nabla_{\theta}\pi_{\theta}(a^{'}|s^{'})Q^{\pi(\theta)}(s^{'},a^{'})\right]\label{eq:pg2}\end{align}\end{figure}
the terms in square brackets in Equation (\ref{eq:pg2}) are simply Equation (\ref{eq:pg_org}) with $a$ and $s^{'}$ replaced by $a^{'}$ and $s^{''}$. Since $\nabla_{\theta}V^{\pi(\theta)}(s^{\infty}) = 0$, Equation (\ref{eq:pg2}) could be written as Equation (\ref{eq:pg3}), where $s_k$ represent the state of $k$ steps after $s$ and $P^{\pi}(s\rightarrow s_k, k)$ already includes integration of intermediate state $s_{k-1},\hdots s_{1}$ before reaching state $s_k$.
\begin{align}
&\nabla_{\theta}V^{\pi(\theta)}(s)\nonumber=\sum_a\nabla_{\theta}\pi_{\theta}(a|s)Q^{\pi(\theta)}(s,a)  + \nonumber \\ 
& \sum_{k=1}\sum_{s_k} \sum_{a_k}\gamma^k P^{\pi}(s\rightarrow s_k, k)\nabla_{\theta}\pi_{\theta}(a_k|s_k)Q^{\pi(\theta)}(s_{k},a_{k})\label{eq:pg3}	
\end{align}
Let objective function with respect to policy be defined to be the value function starting from auxiliary state $s_0$ as in Equation (\ref{eq:obj}).
\begin{align}
J(\pi_{\theta}) = V^{\pi}(s_0)= E_{\pi, \varepsilon}\sum_{t = 0}^{\infty}\gamma^t R_t(S_0 = s)\label{eq:obj}
\end{align} 
The optimal policy could be obtained by gradient accent optimization, leading to the policy gradient algorithm \cite{sutton1998introduction}, as in Equation (\ref{eqn:pgfinal}). 
\begin{align}
&\nabla_{\theta} J(\pi_{\theta})\nonumber=\nabla_{\theta}V^{\pi}(s_0)\nonumber\\
=&\sum_{k=0}\sum_{s_k} \sum_{a_k}\gamma^k P^{\pi}(s_0 \rightarrow s_k, k)\nabla_{\theta}\pi_{\theta}(a_k|s_k)Q^{\pi(\theta)}(s_{k},a_{k})\nonumber\\
=&\sum_{s} \sum_{a}\eta(s)\nabla_{\theta}\pi_{\theta}(a|s)Q^{\pi(\theta)}(s,a)\\
=&\frac{\sum_{s}\eta(s)}{\sum_{s}\eta(s)}\sum_{s} \sum_{a}\eta(s)\nabla_{\theta}\pi_{\theta}(a|s)Q^{\pi(\theta)}(s,a)\\
=&\sum_{\bar{s}}\eta(\bar{s})\sum_{s} \sum_{a}\mu(s)\nabla_{\theta}\pi_{\theta}(a|s)Q^{\pi(\theta)}(s,a)\\
=&\sum_{\bar{s}}\eta(\bar{s})\sum_{s} \sum_{a}\mu(s)\frac{\pi_{\theta}(a|s)}{\pi_{\theta}(a|s)}\nabla_{\theta}\pi_{\theta}(a|s)Q^{\pi(\theta)}(s,a)\label{eq:pg4}\\
\propto&\,E_{\pi}\left[\frac{\nabla_{\theta}\pi_{\theta}(A|S)}{\pi_{\theta}(A|S)} \hat{Q}^{\pi(\theta)}(S,A)\right]
\label{eqn:pgfinal}
\end{align} 
The policy gradient could be augmented to include zero gradient baseline $b(s)$, with respect to objective function $J(\pi_{\theta})$ in Equation (\ref{eq:pg4}), as a function of state $s$, which does not include parameters for policy $\theta$, since $\sum_{a} \nabla_{\theta}\pi_{\theta}(a|s)=0$. To reduce variance of the gradient, the baseline is usually chosen to be the state value function estimator $\hat{V}_w(s)$ to smooth out the variation of $Q(s,a)$ at each state, while $\hat{V}_w(s)$ is updated in a Monte Carlo way by comparing with $\hat{Q}^{\pi_{\theta}}(S,A)=G_t$.

The actor-critic algorithm \cite{sutton1998introduction} decomposes $G_t - V_w(s_t)$ to be $R_t +\gamma V_w(s_{t+1}) - V_w(s_t)$, so bootstrap is used instead of Monte Carlo.

%% file: sec_svi.tex
As a contribution of this paper, we summarize the relationship of evidence $\log p(x)$, KL divergence $D_{KL}$,  cross entropy $H_q(p)$, entropy $H(q)$, free energy $F(\phi, \theta)$ and $ELBO(\phi, \theta)$ in Equation (\ref{eq:vielbo}).
\begin{figure}
\begin{align}
&\log p(x) \nonumber\\=& E_q \left[\log p(x \mid z)\right] - E_q \left[\log \frac{q_{\phi}{(z\mid x)}}{p(z)}\right] + E_q \left[\log \frac{q_{\phi}{(z\mid x)}}{p(z \mid x)}\right]
\nonumber\\=& -D_{KL}(q_{\phi}(z|x)||p(x,z))- E_q \left[\log p(z|x)\right] +\nonumber\\ &E_q \log q_{\phi}(z|x)
\nonumber\\
=&-F(\phi,\theta)+H_q(p) - H(q)
\nonumber\\
=&ELBO(\phi, \theta) +  D_{KL}(q_{\phi}(z \mid x) || p(z \mid x))
\label{eq:vielbo}	
\end{align}
\end{figure}

%% file: sec_pomdp.tex
The latent state distribution associated with belief state can be updated in a Bayesian way in Equation (\ref{eq:belief_pomdp}).
\begin{figure}
	\begin{align}
	&b_{t+1}(s_{t+1}) \nonumber\\
	=& p(s_{t+1}\mid o_t, a_t, b_t)\nonumber\\
	=&\frac{p(s_{t+1}, o_t, a_t, b_t)}{p(o_t, a_t, b_t)}\frac{p(s_{t+1},a_t,b_t)}{p(s_{t+1},a_t,b_t)}\nonumber\\
	=&p(o_t\mid a_t, s_{t+1},b_t)\frac{p(s_{t+1}\mid a_t,b_t)}{p(o_t\mid a_t,b_t)}\\
	=&p(o_t\mid s_{t+1},a_t)\frac{\sum_{{s_t}}p(s_t, s_{t+1}\mid a_t,b_t)}{p(o_t\mid a_t,b_t)}\\
	=&p(o_t\mid s_{t+1},a_t)\frac{\sum_{{s_t}}p(s_{t+1}\mid s_t, a_t,b_t)p(s_{t}\mid a_t,b_t)}{p(o_t\mid a_t,b_t)}\\
	=&p(o_t\mid s_{t+1},a_t)\frac{\sum_{{s_t}}p(s_{t+1}\mid s_t, a_t)p(s_{t}\mid a_t,b_t)}{p(o_t\mid a_t,b_t)}\label{eq:belief_pomdp}
	\end{align}
\end{figure}

%% file: sec_drl.tex
\subsection{Basics of Deep Reinforcement Learning}\label{sec:drl_intro}
Deep Q learning \cite{mnih2015human} makes a breakthrough in using neural network as the functional approximator for value function on complicated tasks. It solves the transition correlation problem by random sampling from a replay memory. Specifically, the reinforcement learning is transformed in a supervised learning task by fitting on the target $R_t + \gamma \underset{a}{max}\,Q(s_{t+1},a)$ from the replay memory with state $s_t$ as input. However, the target can get drifted easily which leads to unstable learning. In \cite{mnih2015human}, a target network is used to provide a stable target for the updating network to be learned before getting updated occasionally.
Double Deep Q learning \cite{van2016deep}, however, solves the problem by having two Q network and update the parameters in a alternating way. 
We review some state of art deep reinforcement learning algorithms from different aspects:
\subsubsection{Off Policy methods} Except for Deep Q Learning \cite{mnih2015human} mentioned above, DDPG \cite{lillicrap2015continuous} extends Deterministic Policy Gradient (DPG) \cite{silver2014deterministic} with deep neural network functional approximator, which is an actor-critic algorithm and works well in continuous action spaces. 
\subsubsection{On Policy methods} A3C \cite{mnih2016asynchronous} stands out in the asynchronous methods in deep learning \cite{mnih2016asynchronous} which can be run in parallel on a single multi-core CPU.
Trust Region  Policy Optimization \cite{schulman2015trust} and Proximal Policy Optimization \cite{schulman2017proximal} assimilates the natural policy gradient, which use a local approximation to the expected return. The local approximation could serve as a lower bound for the expected return, which can be optimized safely subject to the KL divergence constraint between two subsequent policies, while in practice, the constraint is relaxed to be a regularization.
\subsubsection{Goal based Reinforcement Learning} In robot manipulation tasks, the goal could be represented with state in some cases \cite{zhao2019maximum}. Universal Value Function Approximator (UVFA) \cite{schaul2015universal} incorporate the goal into the deep neural network, which let the neural network functional approximator also generalize to goal changes in tasks, similar to Recommendation System \cite{kushwaha2018lesson}. Work of this direction include \cite{andrychowicz2017hindsight, zhao2019maximum}, for example. 
\subsubsection{Replay Memory Manipulation based Method}
Replay memory is a critical component in Deep Reinforcement Learning, which solves the problem of correlated transition in one episode. Beyond the uniform sampling of replay memory in Deep Q Network \cite{mnih2015human},
Prioritized Experience Replay \cite{schaul2015prioritized} improves the performance by giving priority to those transitions with bigger TD error, while Hindsight Experience Replay (HER) \cite{andrychowicz2017hindsight} manipulate the replay memory with changing goals to transition so as to change reward to promote exploration. Maximum entropy regularized multi goal reinforcement learning \cite{zhao2019maximum} gives priority to those rarely occurred trajectory in sampling, which has been shown to improve over HER \cite{zhao2019maximum}. 
\subsubsection{Surrogate policy  optimization} Like surrogate model used in Bayesian Optimization \cite{sun2019high}, lower bound surrogate is also used in Reinforcement Learning. Trust Region Policy Optimization (TRPO) \cite{schulman2015trust} is built on the identity from \cite{kakade2002approximately} in Equation (\ref{eqn:identity}), where $\eta_{\pi^{new}}(s)$ means the state visitation frequency under policy $\pi^{new}$ and advantage $A^{\pi^{old}}(a_t,s_t)=Q^{\pi^{old}}(a_t,s_t) - V^{\pi^{old}}(s_t)$.
\begin{figure}
	\begin{align}
	J(\pi^{new})
	&=J(\pi^{old})  + \sum_s\eta_{\pi^{new}}(s)\sum_a\pi^{new}(a|s)A^{\pi^{old}}(a,s)\label{eqn:identity}
	\end{align}\end{figure}
Based on Policy Advantage \cite{kakade2002approximately} $A_{\pi^{old}, \eta_{old}}(\pi^{new})=\sum_s\eta_{\pi^{old}}(s)\sum_a\pi^{new}(a|s)A^{\pi^{old}}(a,s)$, a local approximation $L_{\pi^{old}}(\pi^{new})$ to Equation (\ref{eqn:identity}) can be defined in Equation (\ref{eqn:policyadv_localapp}), based on which, a surrogate function $M(\pi^{new}, \pi^{old})$ is defined in Equation (\ref{eqn:surrogate_trpo}) that minorizes $J(\pi^{new})$ at $\pi^{old}$, where $D_{KL}^{max}(\pi^{old}, \pi^{new}) =\underset{s}{\max}D_{KL}(\pi^{old}(a|s),\pi^{new}(a|s))$ is the maximum KL divergence, so MM \cite{schulman2015trust} algorithm could be used to improve the policy, leading to the trust region method \cite{schulman2015trust}.
\begin{figure}
	\begin{align}
	&L_{\pi^{old}}(\pi^{new}) &\nonumber\\= &J(\pi^{old})  + \sum_s\eta_{\pi^{old}}(s)\sum_a\pi^{new}(a|s)A^{\pi^{old}}(a_t,s_t)\label{eqn:policyadv_localapp}
	\end{align}
\end{figure}
\begin{figure}
	
	\begin{align}
	&M(\pi^{new},\pi^{old})\nonumber\\ =& L_{\pi^{old}}(\pi^{new})  - \frac{4\underset{a,s}{max}|A^{\pi^{old}}(s,a)| \gamma}{1-\gamma^2}D_{KL}^{max}(\pi^{old}, \pi^{new})\label{eqn:surrogate_trpo}
	\end{align}\end{figure}
\begin{table}[h]\small 
	\caption{Comparison of deep reinforcement learning methods: "S" means state and "A" means action, where "c" means continuous, "d" means discrete. "standalone" means whether the algorithm work independently or needs to be combined with another learning algorithm. "var" means which probability the variational inference is approximating, "p" means whether the method is on policy or off policy. "na" means not applicable}
	\centering
\begin{tabular}{l*{6}{c}r}\hline 
	Algorithm   & S & A &standalone & var & p\\
	\hline
	Deep Q      & c & d  &y & na & off\\
	A3C         & c & c/d &y &na & on\\
	TRPO/PPO    & c & c/d   &y &na &on \\
	DDPG     & c & c   &y &na &off\\
	\hline
	Boltzmann   &d &d &y &na & on\\
	VIME        &c &c  & n &   $p_{\theta}(s_{t+1}|s_{t}, a_t)$ &na\\
	VAST        &c &d  & n & $p(s_t|o_{t-k})$      &na\\
	SoftQ       &c  &c/d  & y & $p(a_t|s_t)$     &on\\
	\hline
\end{tabular}\label{tb:comparision}\end{table}

%% file: sec_undirected_rbm_value_policy.tex
\section{Taxonomy of PGM and VI in deep reinforcement learning}
Despite the success of deep reinforcement learning in many talks, the field still faces some critical challenges. One problem is exploration with sparse reward. In complicated real environment, an agent has to explore for a long trajectory before it can get any reward as feedback. Due to lack of enough rewards, traditional Reinforcement Learning methods performs poorly, which lead to a lot of recent contributions in the exploration methods. 
Another challenge is how to represent policy in extremely large state and action spaces. Furthermore, sometimes it is beneficial to have multimodal behavior for a agent when some trajectory might be equivalent to other trajectories and we want to learn all of them.

In this section, we give detailed explanation on how graphical model and variational inference could be used to model and optimize the reinforcement learning process under these challenges and form a taxonomy of these methods. 

Together with the deep reinforcement learning methods mentioned in section \ref{sec:drl_intro}, we make a comparison of them in Table \ref{tb:comparision}.
\subsection{Policy and value function with undirected graphs}\label{sec:undirected}
\begin{figure}[h!]
	\centering
	\includegraphics[scale=0.6]{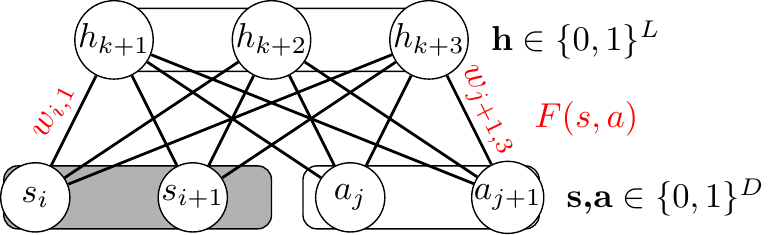}
\caption[short]{Restricted Boltzmann Machine Value and Policy}
\label{fig:PGM_boltzman}
\end{figure}

We first discuss the application of undirected graphs in deep reinforcement learning, which models joint distribution of variables with cliques \cite{bishop2006pattern}. In \cite{sallans2004reinforcement}, the authors use Restricted Boltzmann Machine (RBM) \cite{asja}, which has nice property of tractable factorized posterior distribution over the latent variables conditioned on observed variables. To deal with MDPs of large state and action spaces, they model the state-action value function with the negative free energy of a Restricted Boltzmann Machine. Specifically, the visible states of the Restricted Boltzmann Machine \cite{sallans2004reinforcement} consists of both state $s$ and action $a$ binary variables, as shown in Figure \ref{fig:PGM_boltzman}, where the hidden nodes consist of $L$ binary variables, while state variables $s_i$ are dark colored to represent it can be observed and actions $a_j$ are light colored to represent it need to be sampled. Together with the auxiliary hidden variables, the undirected graph defines a joint probability distribution over state and action pairs, which defines a stochastic policy network that could sample actions out for on policy learning. Since it is pretty easy to calculate the derivative of the free energy $F(s,a)$ with respect to the coefficient $w_{k,j}$ of the network, one could use temporal difference learning to update the coefficients in the network. Thanks to properties of Boltzmann Machine, the conditional distribution of action over state $p(a|s)$, which could be used as a policy, is still Boltzmann distributed as in Equation (\ref{eqn:boltz}), governed by the free energy $F(a,s)$, where $Z(s)$ is the partition function \cite{bishop2006pattern} and the negative free energy to approximate the state action value function $Q(s,a)$. By adjusting the temperature $T$, one could also change between different exploration strength. 
\begin{align}
p(a|s) = {1/Z(s)}e^{-F(s,a)/T} = {1/Z(s)}e^{Q(s,a)/T}\label{eqn:boltz}
\end{align}A few steps of MCMC sampling \cite{bishop2006pattern} could be used to sample actions, as an approximation of the policy, which can be fed into a time difference learning method like SARSA \cite{sutton1998introduction}, to update the state value function $Q(s,a)$'s estimation. Such an on-policy process has been shown to be empirically effective in the large state actions spaces \cite{sallans2004reinforcement}.


%% file: sec_directed_softq.tex
\subsection{Variational Inference on "optimal" Policies}\label{sec:softq}
\subsubsection{policy as "optimal" posterior}
\begin{figure}[]
	\centering
	\includegraphics[scale=0.7]{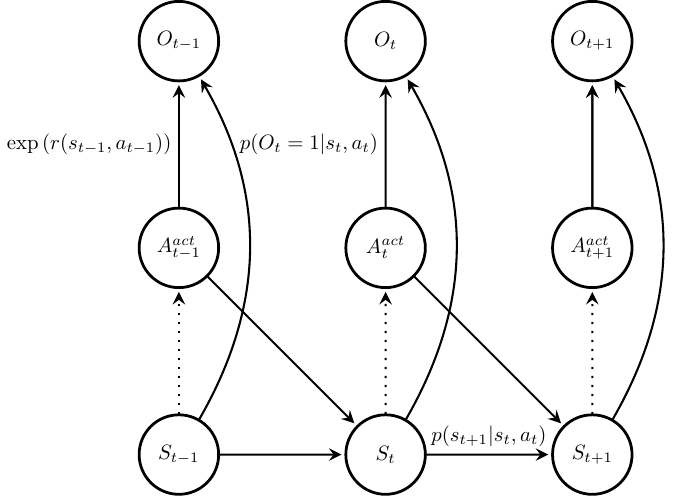}
\caption[short]{Optimal Policy as posterior on actions: $p(a_t|s_t,O_{t:T}=1)$}
\label{fig:pgm_dag_softq}
\end{figure}

The Boltzmann Machine defined Product of Expert Model in \cite{sallans2004reinforcement} works well for large state and action spaces, but are limited to discrete specifically binary state and action variables. For continuous state and action spaces, in \cite{haarnoja2017reinforcement}, the author proposed deep energy based models with Directed Acyclic Graphs (DAG) \cite{bishop2006pattern}, which we re-organize in a different form in Figure \ref{fig:pgm_dag_softq} with annotations added. The difference with respect to  Figure \ref{fig:PGM_MDP} is that, in Figure \ref{fig:pgm_dag_softq}, the reward is not explicit expressed in the directed graphical model. Instead, an auxilliary binary Observable $O$ is used to define whether the corresponding action at the current step is optimal or not. The conditional probability of the action being optimal is $p(O_t = 1|s_t,a_t)=\exp(r(s_t,a_t))$, which connects conditional optimality with the amount of award received by encouraging the agent to take highly rewarded actions in an exponential manner. Note that the reward here must be negative to ensure the validity of probability, which does not hurt generality since reward range can be translated \cite{levine2018reinforcement}.

The Graphical Model in Figure \ref{fig:pgm_dag_softq} in total defines the trajectory likelihood or the evidence in Equation (\ref{eqn:softq_trajectory_likelihood}): 
\begin{align}
p(\tau) 
&= \left[p(s_1)\prod_t p(s_{t+1}|s_t,a_t)\right]\exp\left(\sum_t r(s_t,a_t)\right)\label{eqn:softq_trajectory_likelihood}
\end{align}.

By doing so, the author is forcing a form of functional expression on top of the conditional independence structure of the graph by assigning a likelihood. In this way, calculating the optimal policy of actions distributions becomes an inference problem of calculating the posterior $p(a_t|s_t,O_{t:T}=1)$, which reads as, conditional on optimality from current time step until end of episode, and the current current state to be $s_t$, the distribution of action $a_t$, and this posterior corresponds to the optimal policy. Observing the d-separation  from Figure \ref{fig:pgm_dag_softq}, $O_{1:{t-1}}$ is conditionally independent of $a_t$ given $s_t$, $(O_{1:{t-1}} \bigCI A^{act}_t)\mid {S_t}$, so  $p(a_t|s_t, O_{1:t-1}, O_{t:T}) = p(a_t|s_t, O_{t:T})$

\subsubsection{Message passing for exact inference on the posterior}
In this section, we give detailed derivation on conducting exact inference on the policy posterior which is not given in \cite{levine2018reinforcement}. 
Similar to the forward-backward message passing algorithm \cite{bishop2006pattern} in Hidden Markov Models \cite{bishop2006pattern}, the posterior $p(a_t|s_t,O_{t:T}=1)$ could also be calculated by passing messages. We offer a detailed derivation of the decomposition of the posterior $p(a_t|s_t,O_{t:T}=1)$ in Equation (\ref{eqn:betadef}), which is not available in \cite{levine2018reinforcement}. 
\begin{figure}
\begin{align}
&p(a_t|s_t,O_{t:T}=1)\nonumber\\
=&\frac{p(a_t,s_t,O_{t:T}=1)}{p(s_t,O_{t:T}=1)}\nonumber\\
=&\frac{p(O_{t:T}=1|a_t,s_t)p(a_t, s_t)}{p(s_t,O_{t:T}=1)} \nonumber\\
=&\frac{p(O_{t:T}=1|a_t,s_t)p(a_t| s_t)p(s_t)}{\int _{a_t{'}}p(s_t,a_t^{'},O_{t:T}=1)d\{a_t^{'}\}}\nonumber\\
=&\frac{p(O_{t:T}=1|a_t,s_t)p(a_t| s_t)p(s_t)}{\int _{a_t^{'}}p(O_{t:T}=1|a_t{'},s_t)p(a_t{'}| s_t)p(s_t)d\{a_t^{'}\}}\nonumber\\
=&\frac{p(O_{t:T}=1|a_t,s_t)p(a_t| s_t)}{\int _{a_t^{'}}p(O_{t:T}=1|a_t{'},s_t)p(a_t{'}| s_t)d\{a_t^{'}\}}\nonumber\\
=&\frac{\beta(a_t,s_t)}{\int_{a_t^{'}}\beta(a_t^{'},s_t)d\{a_t^{'}\}}\nonumber\\
=&\frac{\beta(a_t,s_t)}{\beta(s_t)}\label{eqn:betadef}
\end{align}
\end{figure}
In Equation (\ref{eqn:betadef}), we define 	message $\beta(a_t,s_t) = p(O_{t:T}=1|a_t,s_t)p(a_t| s_t)$
and message $\beta(s_t) = \int_{a_t^{'}}\beta(a_t^{'},s_t)d\{a_t^{'}\}$. If we consider $p(a_t|s_t)$ as a prior with a trivial form of uniform distribution \cite{levine2018reinforcement}, the only policy related term becomes $p(O_{t:T}=1|a_t,s_t)$.

In contrast to HMM, here, only the backward messages are relevant. Additionally, the backward message $\beta(a_t, s_t)$ here  is not a probability distribution as in HMM, instead, is just a probability. In Figure \ref{fig:pgm_dag_softq}, the backward message $\beta(a_t, s_t)$ could be decomposed recursively. Since in \cite{levine2018reinforcement} the author only give the conclusion without derivation, we give a detailed derivaion of this recursion in Equation (\ref{eqn:backmessage}).
\begin{figure}
\begin{align}
&\beta(s_t, a_t) \nonumber\\
=& p(O_t =1, O_{t+1:T}=1|s_t,a_t)\nonumber\\
=&\frac{\int p(O_t =1, O_{t+1:T}=1, s_t,a_t, s_{t+1}, a_{t+1})d\{s_{t+1}, a_{t+1}\}}{p(s_t, a_t)}\nonumber\\
=&\int p(O_{t+1:T}=1,s_{t+1}, a_{t+1},O_t=1|s_t,a_t)d\{s_{t+1}, a_{t+1}\}\nonumber\\ 
=&\int p(O_{t+1:T}=1,s_{t+1}, a_{t+1}|s_t,a_t)p(O_t=1|s_t,a_t)\nonumber\\&d\{s_{t+1}, a_{t+1}\} \tag{$(O_{t+1:T},S_{t+1}, A_{t+1} \bigCI O_t) \mid S_t,A_t$} \\  
=&\int \frac{p(O_{t+1:T}=1,s_{t+1}, a_{t+1})}{p(s_{t+1}, a_{t+1})}\frac{p(s_{t+1},s_t,a_t)}{p(s_t,a_t)}\nonumber\\&p(O_t=1|s_t, a_t)d\{s_{t+1}, a_{t+1}\}\nonumber\\
=&\int p(O_{t+1:T}=1|s_{t+1}, a_{t+1})p(s_{t+1}|s_t,a_t)p(O_t=1|s_t,a_t)\nonumber\\&d\{s_{t+1}, a_{t+1}\}\nonumber\\ 
=&\int \beta(s_{t+1})p(s_{t+1}|s_t,a_t)p(O_t=1|s_t,a_t)d s_{t+1}
\label{eqn:backmessage}
\end{align}
\end{figure}
The recursion in Equation (\ref{eqn:backmessage}) start from the last time point $T$ of an episode.
\subsubsection{Connection between Message Passing and Bellman equation}
If we define Q function in Equation (\ref{eqn:qbeta}) and V function in Equation (\ref{eqn:vbeta})
\begin{equation}
Q(s_t,a_t) = \log(\beta(a_t,s_t))\label{eqn:qbeta}\end{equation} 
\begin{figure}
\begin{align}
V(s_t)&=\log \beta(s_t)=\log \int \beta(s_t,a_t)da_t\nonumber\\
&= \log \int exp(Q(s_t,a_t))da_t\approx \underset{a_t}{max}Q(s_t,a_t)\label{eqn:vbeta}\end{align}
\end{figure}

then the corresponding policy could be written as Equation (\ref{eqn:softpolicy}).
\begin{equation}
\pi(a_t|s_t) = p(a_t|s_t, O_{t:T}=1) = \exp(Q(s_t,a_t) - V(s_t))\label{eqn:softpolicy}\end{equation}
Taking the logrithm of Equation (\ref{eqn:backmessage}), we get Equation (\ref{eqn:messagebellman})
\begin{figure}	
\begin{align}
	&\log(\beta(s_t,a_t))\nonumber\\
	&=\log \int \beta(s_{t+1})p(s_{t+1}|s_t,a_t)p(O_t=1|s_t,a_t)d s_{t+1}\nonumber\\
	&=\log \int \exp[r(s_t,a_t) + V(s_{t+1})]p(s_{t+1}|s_t,a_t)d s_{t+1}\nonumber\\
	&=r(s_t,a_t) +\log \int \exp( V(s_{t+1}))p(s_{t+1}|s_t,a_t)d s_{t+1}\label{eqn:messagebellman}
\end{align}
\end{figure}
which reduces to the risk seeking backup in Equation (\ref{eqn:risk_bellman}) as mentioned in \cite{levine2018reinforcement}:
\begin{equation}
	Q(s_t,a_t) = r(s_t, a_t) + \log E_{s_{t+1}\sim p(s_{t+1}|s_t,a_t)}[\exp(V(s_{t+1}))]\label{eqn:risk_bellman}
\end{equation}

The mathematical insight here is that if we define the messages passed on the Directed Acyclic Graph in Figure \ref{fig:pgm_dag_softq}, then message passing correspond to a peculiar version Bellman Equation like backup, which lead to an unwanted risk seeking behavior \cite{levine2018reinforcement}: when compared to Equation (\ref{eqn:v2q}), the Q function here is taking a softmax instead of expectation over the next state.

\subsubsection{Variational approximation to "optimal" policy}
Since the exact inference lead to unexpected behavior, approximate inference could be used. 
The optimization of the policy could be considered as a variational inference problem, and
we use the variational policy of the action posterior distribution $q(a_t|s_t)$, which could be represented by a neural network,  to compose the proposal variational likelihood of the trajectory as in Equation (\ref{eqn:softq_vardistribution}):
\begin{figure}
	\begin{align}
	q(\tau) &= 
	p(s_1)\prod_t [p(s_{t+1}|s_t,a_t)q(a_t|s_t)] \label{eqn:softq_vardistribution}
	\end{align}
\end{figure}
where the initial state distribution $p(s_1)$ and the environmental dynamics of state transmission is kept intact. Using the proposal trajectory as a pivot, we could derive the Evidence Lower Bound (ELBO) of the optimal trajectory as in Equation (\ref{eqn:softq_elbo}), which correspond to an interesting objective function of reward plus entropy return, as in Equation (\ref{eqn:soft_elbo_obj}).

\input{derivation_soft_q}
\subsubsection{Examples}
In \cite{haarnoja2017reinforcement}, the state action value function is defined in Equation (\ref{eqn:defsoftq}).
\begin{figure}
\begin{equation}
Q^{\pi}_{soft}(s,a)= r_0 + E_{r\sim\pi, s_0=s,a_0=a}[\sum_{t=1}^{\infty}\gamma^t(r_t + \alpha H(\pi(.|s_t)))]\label{eqn:defsoftq}
\end{equation}\end{figure}
and a soft version of Bellman update similar to  Q Learning \cite{sutton1998introduction} is carried out, which lead to policy improvement with respect to the corresponding functional objective in Equation (\ref{eqn:softq_jpi}).
\begin{figure}
\begin{align}
&J(\pi) \nonumber \\=& \sum_t E_{(s_t, a_t)\sim\rho_{\pi}}\sum_{l=t}^{\infty}\gamma^{l-t}E_{(s_l,a_l)}[r(s_l,a_l)+ \nonumber\\& 
\alpha H(\pi(.|s_l))|s_t,a_t]]\nonumber\\
=&\sum_t E_{(s_t, a_t)\sim\rho_{\pi}}[Q_{soft}^{\pi}(s_t,a_t) + \alpha H(\pi(.|s_t))]\label{eqn:softq_jpi}
\end{align}
\end{figure}
Setting policy as Equation (\ref{eqn:softpolicy}) lead to policy improvement.
We offer a detailed proof for a key formula in Equation (\ref{eqn:maxentpolicyimprove}), which is stated in Equation (19) of \cite{haarnoja2017reinforcement} without proof. In Equation (\ref{eqn:maxentpolicyimprove}), we use $\pi(\cdot|s)$ to implicitly represent $\pi(a|s)$ to avoid symbol aliasing whenever necessary.
\begin{figure}
\begin{align}
&H(\pi(\cdot|s)) + E_{a\sim \pi}[Q_{soft}^{\pi}(s,a)] \nonumber\\
=& - \int_a \pi(a|s)[\log\pi(a|s) - Q_{soft}^{\pi}(s,a)]da \nonumber\\
=& - \int_a \pi(a|s)[\log\pi(a|s) - \log [\exp (Q_{soft}^{\pi}(s,a))]]da \nonumber\\
=& - \int_a \pi(a|s)[\log\pi(a|s) - \log [\frac{\exp (Q_{soft}^{\pi}(s,a))}{\int \exp (Q_{soft}^{\pi}(s,a^{'}))da^{'}}\nonumber\\&\int \exp (Q_{soft}^{\pi}(s,a^{'}))da^{'}]]da \nonumber\\
=& - \int_a \pi(a|s)[\log\pi(a|s) - \log [\Tilde{\pi}(a|s) ]- \nonumber\\&log \int exp (Q_{soft}^{\pi}(s,a^{'}))]da^{'} \nonumber \\
=& -D_{KL}(\pi(\cdot|s)||\Tilde{\pi}(\cdot|s)) +  \log \int \exp (Q_{soft}^{\pi}(s,a^{'}))da^{'}
\label{eqn:maxentpolicyimprove}
\end{align}
\end{figure}
For the rest of the proof, we invite the reader to read the appendix of \cite{haarnoja2017reinforcement}.
Algorithms of the this kind of maximum entropy family also include Soft Actor Critic \cite{haarnoja2018soft}.

%% file: derivation_soft_q.tex

\begin{figure}
\begin{align}
&\log(p(O_{1:T})) \nonumber\\
=&\log \int p(O_{1:T}=1,s_{1:T},a_{1:T})\frac{q(s_{1:T}, a_{1:T})}{q(s_{1:T}, a_{1:T})}ds_{1:T}da_{1:T} \nonumber\\
=& \log E_{q(s_{1:T}, a_{1:T})} \frac{p(O_{1:T}=1,s_{1:T},a_{1:T})}{q(s_{1:T}, a_{1:T})} \nonumber\\
\geq&  E_{q(s_{1:T}, a_{1:T})} [\log p(O_{1:T}=1,s_{1:T},a_{1:T})- \log q(s_{1:T}, a_{1:T})]
\label{eqn:softq_elbo}\\
=& -D_{KL}(q(\tau)|p(\tau)) \\
=& E_{q(s_{1:T}, a_{1:T})}[\sum_{t=1:T} [r(s_t,a_t) - \log q(a_t|s_t)]] \nonumber\\
=& \sum_{t=1:T}E_{s_t,a_t}[r(s_t,a_t) + H(\pi(a_t|s_t))]\label{eqn:soft_elbo_obj} 
\end{align}\end{figure}


%% file: sec_vi_env.tex
\subsection{Variational Inference on the Environment}\label{sec:var_on_env}
Another direction of using Variational Inference in Reinforcement Learning is to learn an environmental model, either on the dynamics or the latent state space posterior.
\subsubsection{Variational inference on transition model}
\begin{figure}[]
	\centering
	\includegraphics[scale=0.6]{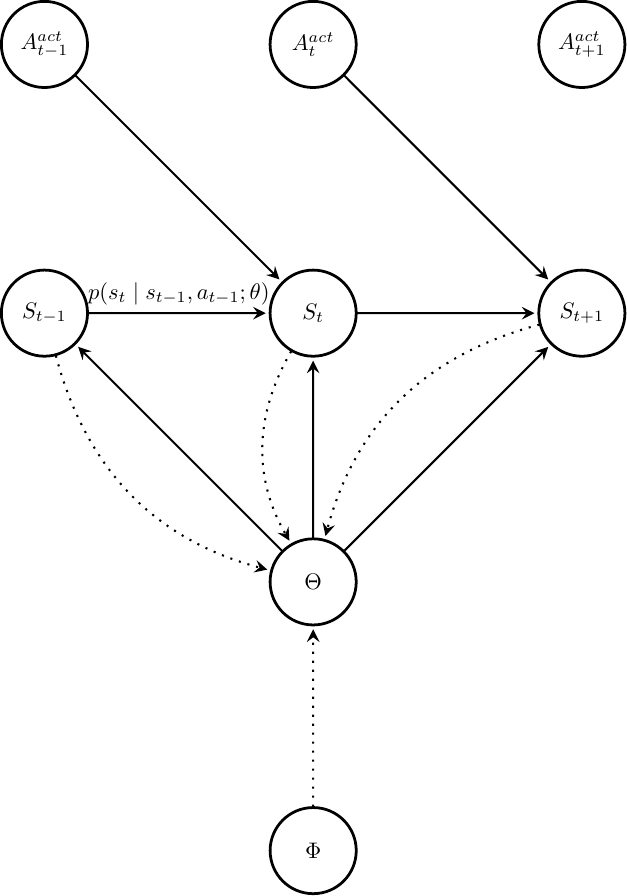}
	\caption[short]{Probabilistic Graphical Model For VIME}
	\label{fig:PGM_VIME}
\end{figure}

In Variational Information Maximizing Exploration (VIME) \cite{houthooft2016vime}, where dynamic model $p_{\theta}(s_{t+1}|s_t,a_t)$ for the agent's interaction with the environment is modeled using Bayesian Neural Network \cite{blundell2015weight}. The R.V. for $\theta$ is denoted by $\Theta$, and is treated in a Bayesian way by modeling the weight $\theta$ uncertainty of a neural network. We represent this model with the graphical model in Figure \ref{fig:PGM_VIME}, which is not given in \cite{houthooft2016vime}. The belief uncertainty about the environment is modeled as entropy of the  posterior distribution of the neural network weights $H(\Theta|\xi_t)$ based on trajectory   observations $\xi_t = \{s_{1:t},a_{1:t-1}\}$. The method encourages taking exploratory actions by alleviating the average information gain of the agent's belief about the environment after observing a new state $s_{t+1}$, which is $E_{p(s_{t+1}|\xi_t,a_t)}D_{KL}(p(\theta|\xi_{t+1})||p(\theta|\xi_{t}))$, and this is equivalent to the entropy minus conditional entropy $H(\Theta|\xi_t,a_t)-H(\Theta|\xi_t,a_t, s_{t+1})=H(\Theta|\xi_t,a_t)-H(\Theta|\xi_{t+1})$.
With the help of Equation (\ref{eqn:kl_dyn}), as derived following the definition of conditional mutual information, we derive in Equation (\ref{eqn:ent_mutualinfo}) that the conditional entropy difference is actually the average information gain, which is equal to the conditional mutual information $I(\Theta, S_{t+1}|\xi_t, a_t)$ between environmental parameter $\Theta$ and the new state $S_{t+1}$. 
Such a derivation is not given in \cite{houthooft2016vime}.
\newcommand{\kld}[2]{\ensuremath{D_{KL}\infdivx{#1}{#2}}\xspace}
\begin{figure}
\begin{align}
I(X;Y\mid Z)&=\int_{x,y,z}p(z)p(x,y|z)\log \frac{p(x,y|z)}{p(x|z)p(y|z)}dxdydz\nonumber\\
&=-\int_{x,y,z}p(z)p(x,y|z)\log p(x|z)dxdzdy +\nonumber\\
&+\int_{x,y,z}p(x,y,z)\log p(x|y,z)dxdzdy\nonumber\\
&=H(X\mid Z) - H(X\mid Y,Z)\label{eqn:kl_dyn}
\end{align}\end{figure}
\begin{figure}
\begin{align}
&H(\Theta|\xi_t, a_t)-H(\Theta|\xi_{t}, a_t, s_{t+1})=I(\Theta, S_{t+1}|\xi_t, a_t)\nonumber\\
=&E_{\xi_{t},a_t}\int_{\Theta,\mathcal{S}}p(s_{t+1},\theta|\xi_{t},a_t)\log[\frac{p(s_{t+1},\theta|\xi_{t},a_t)}{p(\theta|\xi_t)p(s_{t+1}|\xi_t,a_t)}]d\theta \nonumber\\&ds_{t+1}\nonumber\\
=&E_{\xi_{t},a_t}\int_{\Theta,\mathcal{S}}p(s_{t+1}|\xi_{t},a_t)p(\theta|\xi_{t+1})\log[\frac{p(\theta|\xi_{t+1})}{p(\theta|\xi_t)}]d\theta ds_{t+1}\nonumber\\
=&E_{\xi_{t},a_t}E_{p(s_{t+1}|\xi_t,a_t)}D_{KL}(p(\theta|\xi_{t+1})||p(\theta|\xi_{t}))
\label{eqn:ent_mutualinfo}
\end{align}\end{figure}
Based on Equation (\ref{eqn:ent_mutualinfo}), an intrinsic reward can be augmented from the environmental reward function, thus the method could be incorporated with any existing reinforcement learning algorithms for exploration, TRPO \cite{schulman2015trust}, for example.
Upon additional observation of action $a_t$ and state $s_{t+1}$ pair on top of trajectory history $\xi_t$, the posterior on the distribution of the environmental parameter $\theta$, $p(\theta|\xi_t)$, could be updated to be $p(\theta|\xi_{t+1})$ in a Bayesian way as derived in Equation (\ref{eqn:bayesupdate_env_dynamics}), which is first proposed in \cite{sun2011planning}.
\begin{figure}
\begin{align}
p(\theta|\xi_{t+1})
=&\frac{p(\theta,\xi_{t}, a_t, s_{t+1})}{p(\xi_{t}, a_t, s_{t+1})}\nonumber\\
=&\frac{p(s_{t+1}|\theta,\xi_{t}, a_t)p(\theta,\xi_{t}, a_t)}{p(\xi_{t}, a_t, s_{t+1})}\nonumber\\
=&\frac{p(s_{t+1}|\theta,\xi_{t}, a_t)p(\theta,\xi_{t}, a_t)}{p(a_t, \xi_t)p(s_{t+1}|a_t, \xi_{t})}\nonumber\\
=&\frac{p(s_{t+1}|\theta,\xi_{t}, a_t)p(\theta|\xi_{t}, a_t)}{p(s_{t+1}|a_t, \xi_{t})}\nonumber\\
=&\frac{p(s_{t+1}|\theta,\xi_{t}, a_t)p(\theta|\xi_{t})}{p(s_{t+1}|a_t ,\xi_{t})}\label{eqn:bayesupdate_env_dynamics}
\end{align}\end{figure}
In Equation (\ref{eqn:bayesupdate_env_dynamics}), the denominator can be written as Equation (\ref{eqn:dyn_nn}), so that the dynamics of the environment modeled by neural network weights $\theta$, $p(s_{t+1}|\theta,a_t, \xi_{t})$, could be used.
\begin{figure}
\begin{align}
&p(s_{t+1}|a_t, \xi_{t})\nonumber\\
&= \int_{\Theta}p(s_{t+1},\theta|a_t, \xi_{t})d\theta\nonumber\\
&=\int_{\Theta}\frac{p(s_{t+1},\theta,a_t, \xi_{t})}{p(a_t, \xi_{t})}d\theta\nonumber\\
&=\int_{\Theta}\frac{p(s_{t+1}|\theta,a_t, \xi_{t})p(\theta,a_t, \xi_{t})}{p(a_t, \xi_{t})}d\theta\nonumber\\
&=\int_{\Theta}p(s_{t+1}|\theta,a_t, \xi_{t})p(\theta|\xi_t)d\theta \label{eqn:dyn_nn}
\end{align}\end{figure}
The last step of Equation (\ref{eqn:dyn_nn}) makes use of $p(\theta|\xi_t,a_t) = p(\theta|\xi_t)$.

Since the integral in Equation (\ref{eqn:dyn_nn}) is not tractable, variational treatment over the neural network weights posterior distribution $p(\theta|\xi_{t})$ is used, characterized by variational parameter $\phi$, as shown in the dotted line in Figure \ref{fig:PGM_VIME}. The variational posterior about the model parameter $\theta$, updated at each step, could than be used to calculate the intrinsic reward in Equation (\ref{eqn:ent_mutualinfo}). 
\subsubsection{Variational Inference on hidden state posterior}
\begin{figure}[]
	\centering
	\includegraphics[scale=0.7]{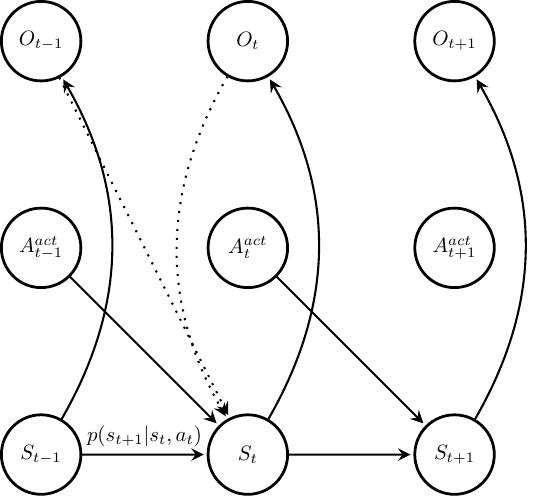}
\caption[short]{Graphical Model for Variation State Tabulation}
\label{fig:pgm_dag_vast}
\end{figure}

In Variational State Tabulation (VaST) \cite{corneil2018efficient}, the author assume the high dimensional observed state to be represented by Observable $O$, while the transition happens at the latent state space represented by $S$, which is finite and discrete. The author assume a factorized form of observation and latent space joint probability, which we explicitly state in Equation (\ref{eqn:clikelihood_vast}).
\begin{align}
p(O,S)=\pi_{\theta_0}(s_0)\prod_{t=0}^{T}p_{\theta^R}(o_t|s_t)\prod_{t=1}^{T}p_{\theta^T}(s_{t}|s_{t-1},a_{t-1})\label{eqn:clikelihood_vast}\end{align}
Additionally, we characterize Equation (\ref{eqn:clikelihood_vast}) with the probabilistic graphical model in Figure \ref{fig:pgm_dag_vast} which does not exist in \cite{corneil2018efficient}. Compared to Figure \ref{fig:pgm_dag_softq}, here the latent state $S$ is in discrete space instead of high dimension, and the observation is a high dimensional image instead of binary variable to indicate optimal action. By assuming a factorized form of the variational posterior in Equation (\ref{eqn:vast_posterior}),
\begin{align}
q(S_{0:T}|O_{0:T}) = \prod_{t=0}^{T}q_{\phi}(S_{t}|O_{t-k:t})\label{eqn:vast_posterior}
\end{align}The author assume the episode length to be $T$, and default frame prior observation to be blank frames. The Evidence Lower Bound (ELBO) of the observed trajectory of Equation (\ref{eqn:clikelihood_vast}) could be easily represented by a Varitional AutoEncoder \cite{doersch2016tutorial} like architecture, where the encoder $q_{\phi}$, together with the reparametrization trick \cite{doersch2016tutorial}, maps the observed state $O$ into parameters for the Con-crete distribution \cite{maddison2016concrete}, so backprobagation could be used on deterministic variables to update the weight of the network based on the ELBO, which is decomposed into different parts of the reconstruction losses of the variational autoencoder like architecture. Like VIME \cite{houthooft2016vime}, VaSt could be combined with other reinforcement learning algorithms. Here prioritized sweeping \cite{sutton1998introduction} is carried out on the Heviside activation of the encoder output directly, by counting the transition frequency, instead of waiting for the slowly learned environmental transition model $p_{\theta^T}(s_t|s_{t-1},a_{t-1})$ in Equation (\ref{eqn:clikelihood_vast}). A potential problem of doing so is aliasing between latent state $s$ and observed state $o$. To alleviate this problem, in \cite{corneil2018efficient}, the author actively relabel the transition history in the replay memory once found the observable has been assigned a different latent discrete state. 

%% file: main_pgm_avi_rl_ieee_bare_conf.bbl
\begin{thebibliography}{10}
\providecommand{\url}[1]{#1}
\csname url@samestyle\endcsname
\providecommand{\newblock}{\relax}
\providecommand{\bibinfo}[2]{#2}
\providecommand{\BIBentrySTDinterwordspacing}{\spaceskip=0pt\relax}
\providecommand{\BIBentryALTinterwordstretchfactor}{4}
\providecommand{\BIBentryALTinterwordspacing}{\spaceskip=\fontdimen2\font plus
\BIBentryALTinterwordstretchfactor\fontdimen3\font minus
  \fontdimen4\font\relax}
\providecommand{\BIBforeignlanguage}[2]{{%
\expandafter\ifx\csname l@#1\endcsname\relax
\typeout{** WARNING: IEEEtran.bst: No hyphenation pattern has been}%
\typeout{** loaded for the language `#1'. Using the pattern for}%
\typeout{** the default language instead.}%
\else
\language=\csname l@#1\endcsname
\fi
#2}}
\providecommand{\BIBdecl}{\relax}
\BIBdecl

\bibitem{mnih2015human}
V.~Mnih, K.~Kavukcuoglu, D.~Silver, A.~A. Rusu, J.~Veness, M.~G. Bellemare,
  A.~Graves, M.~Riedmiller, A.~K. Fidjeland, G.~Ostrovski \emph{et~al.},
  ``Human-level control through deep reinforcement learning,'' \emph{Nature},
  vol. 518, no. 7540, p. 529, 2015.

\bibitem{schulman2015trust}
J.~Schulman, S.~Levine, P.~Abbeel, M.~Jordan, and P.~Moritz, ``Trust region
  policy optimization,'' in \emph{International conference on machine
  learning}, 2015, pp. 1889--1897.

\bibitem{sun2019reinbo}
X.~Sun, J.~Lin, and B.~Bischl, ``Reinbo: Machine learning pipeline search and
  configuration with bayesian optimization embedded reinforcement learning,''
  2019.

\bibitem{houthooft2016vime}
R.~Houthooft, X.~Chen, Y.~Duan, J.~Schulman, F.~De~Turck, and P.~Abbeel,
  ``Vime: Variational information maximizing exploration,'' in \emph{Advances
  in Neural Information Processing Systems}, 2016, pp. 1109--1117.

\bibitem{corneil2018efficient}
D.~Corneil, W.~Gerstner, and J.~Brea, ``Efficient model-based deep
  reinforcement learning with variational state tabulation,'' \emph{arXiv
  preprint arXiv:1802.04325}, 2018.

\bibitem{blei2017variational}
D.~M. Blei, A.~Kucukelbir, and J.~D. McAuliffe, ``Variational inference: A
  review for statisticians,'' \emph{Journal of the American Statistical
  Association}, vol. 112, no. 518, pp. 859--877, 2017.

\bibitem{bishop2006pattern}
C.~M. Bishop, \emph{Pattern recognition and machine learning}.\hskip 1em plus
  0.5em minus 0.4em\relax springer, 2006.

\bibitem{asja}
A.~Fischer and C.~Igel, ``Training restricted boltzmann machines: An
  introduction,'' \emph{Pattern Recognition}, vol.~47, pp. 25--39, 01 2014.

\bibitem{sun2019resampling}
X.~Sun, A.~Gossmann, Y.~Wang, and B.~Bischl, ``Variational resampling based
  assessment of deep neural networks under distribution shift,'' 2019.

\bibitem{blundell2015weight}
C.~Blundell, J.~Cornebise, K.~Kavukcuoglu, and D.~Wierstra, ``Weight
  uncertainty in neural networks,'' \emph{arXiv preprint arXiv:1505.05424},
  2015.

\bibitem{kingma2013auto}
D.~P. Kingma and M.~Welling, ``Auto-encoding variational bayes,'' \emph{arXiv
  preprint arXiv:1312.6114}, 2013.

\bibitem{sutton1998introduction}
R.~S. Sutton, A.~G. Barto \emph{et~al.}, \emph{Introduction to reinforcement
  learning}.\hskip 1em plus 0.5em minus 0.4em\relax MIT press Cambridge, 1998,
  vol.~2, no.~4.

\bibitem{levine2018reinforcement}
S.~Levine, ``Reinforcement learning and control as probabilistic inference:
  Tutorial and review,'' \emph{arXiv preprint arXiv:1805.00909}, 2018.

\bibitem{zhao2019maximum}
R.~Zhao, X.~Sun, and V.~Tresp, ``Maximum entropy-regularized multi-goal
  reinforcement learning,'' \emph{arXiv preprint arXiv:1905.08786}, 2019.

\bibitem{kaelbling1998planning}
L.~P. Kaelbling, M.~L. Littman, and A.~R. Cassandra, ``Planning and acting in
  partially observable stochastic domains,'' \emph{Artificial i ntelligence},
  vol. 101, no. 1-2, pp. 99--134, 1998.

\bibitem{van2016deep}
H.~Van~Hasselt, A.~Guez, and D.~Silver, ``Deep reinforcement learning with
  double q-learning,'' in \emph{Thirtieth AAAI conference on artificial
  intelligence}, 2016.

\bibitem{lillicrap2015continuous}
T.~P. Lillicrap, J.~J. Hunt, A.~Pritzel, N.~Heess, T.~Erez, Y.~Tassa,
  D.~Silver, and D.~Wierstra, ``Continuous control with deep reinforcement
  learning,'' \emph{arXiv preprint arXiv:1509.02971}, 2015.

\bibitem{silver2014deterministic}
D.~Silver, G.~Lever, N.~Heess, T.~Degris, D.~Wierstra, and M.~Riedmiller,
  ``Deterministic policy gradient algorithms,'' 2014.

\bibitem{mnih2016asynchronous}
V.~Mnih, A.~P. Badia, M.~Mirza, A.~Graves, T.~Lillicrap, T.~Harley, D.~Silver,
  and K.~Kavukcuoglu, ``Asynchronous methods for deep reinforcement learning,''
  in \emph{International conference on machine learning}, 2016, pp. 1928--1937.

\bibitem{schulman2017proximal}
J.~Schulman, F.~Wolski, P.~Dhariwal, A.~Radford, and O.~Klimov, ``Proximal
  policy optimization algorithms,'' \emph{arXiv preprint arXiv:1707.06347},
  2017.

\bibitem{schaul2015universal}
T.~Schaul, D.~Horgan, K.~Gregor, and D.~Silver, ``Universal value function
  approximators,'' in \emph{International Conference on Machine Learning},
  2015, pp. 1312--1320.

\bibitem{kushwaha2018lesson}
N.~Kushwaha, X.~Sun, B.~Singh, and O.~Vyas, ``A lesson learned from pmf based
  approach for semantic recommender system,'' \emph{Journal of Intelligent
  Information Systems}, vol.~50, no.~3, pp. 441--453, 2018.

\bibitem{andrychowicz2017hindsight}
M.~Andrychowicz, F.~Wolski, A.~Ray, J.~Schneider, R.~Fong, P.~Welinder,
  B.~McGrew, J.~Tobin, O.~P. Abbeel, and W.~Zaremba, ``Hindsight experience
  replay,'' in \emph{Advances in Neural Information Processing Systems}, 2017,
  pp. 5048--5058.

\bibitem{schaul2015prioritized}
T.~Schaul, J.~Quan, I.~Antonoglou, and D.~Silver, ``Prioritized experience
  replay,'' \emph{arXiv preprint arXiv:1511.05952}, 2015.

\bibitem{sun2019high}
X.~Sun, A.~Bommert, F.~Pfisterer, J.~Rahnenf{\"u}hrer, M.~Lang, and B.~Bischl,
  ``High dimensional restrictive federated model selection with multi-objective
  bayesian optimization over shifted distributions,'' \emph{arXiv preprint
  arXiv:1902.08999}, 2019.

\bibitem{kakade2002approximately}
S.~Kakade and J.~Langford, ``Approximately optimal approximate reinforcement
  learning,'' in \emph{ICML}, vol.~2, 2002, pp. 267--274.

\bibitem{sallans2004reinforcement}
B.~Sallans and G.~E. Hinton, ``Reinforcement learning with factored states and
  actions,'' \emph{Journal of Machine Learning Research}, vol.~5, no. Aug, pp.
  1063--1088, 2004.

\bibitem{haarnoja2017reinforcement}
T.~Haarnoja, H.~Tang, P.~Abbeel, and S.~Levine, ``Reinforcement learning with
  deep energy-based policies,'' in \emph{Proceedings of the 34th International
  Conference on Machine Learning-Volume 70}.\hskip 1em plus 0.5em minus
  0.4em\relax JMLR. org, 2017, pp. 1352--1361.

\bibitem{haarnoja2018soft}
T.~Haarnoja, A.~Zhou, P.~Abbeel, and S.~Levine, ``Soft actor-critic: Off-policy
  maximum entropy deep reinforcement learning with a stochastic actor,''
  \emph{arXiv preprint arXiv:1801.01290}, 2018.

\bibitem{sun2011planning}
Y.~Sun, F.~Gomez, and J.~Schmidhuber, ``Planning to be surprised: Optimal
  bayesian exploration in dynamic environments,'' in \emph{International
  Conference on Artificial General Intelligence}.\hskip 1em plus 0.5em minus
  0.4em\relax Springer, 2011, pp. 41--51.

\bibitem{doersch2016tutorial}
C.~Doersch, ``Tutorial on variational autoencoders,'' \emph{arXiv preprint
  arXiv:1606.05908}, 2016.

\bibitem{maddison2016concrete}
C.~J. Maddison, A.~Mnih, and Y.~W. Teh, ``The concrete distribution: A
  continuous relaxation of discrete random variables,'' \emph{arXiv preprint
  arXiv:1611.00712}, 2016.

\end{thebibliography}
